\documentclass[acmtog]{acmart}

\renewcommand\footnotetextcopyrightpermission[1]{} 
\makeatletter
\renewcommand\@formatdoi[1]{\ignorespaces}
\makeatother

\usepackage{algpseudocode}
\usepackage{algorithm}
\usepackage{stackrel}
\usepackage{dblfloatfix}

\graphicspath{{"./"}}

\begin{document}

\title{Non-line-of-sight Imaging with Partial Occluders and Surface Normals} 

\author{Felix Heide}
\affiliation{%
  \institution{Stanford University}
}
\author{Matthew O'Toole}
\affiliation{%
  \institution{Stanford University}
}
\author{Kai Zang}
\affiliation{%
  \institution{Stanford University}
}
\author{David Lindell}
\affiliation{%
  \institution{Stanford University}
}
\author{Steven Diamond}
\affiliation{%
  \institution{Stanford University}
}
\author{Gordon Wetzstein}
\affiliation{%
  \institution{Stanford University}
}

\renewcommand\shortauthors{Heide et al.}

\begin{abstract}
Imaging objects obscured by occluders is a significant challenge for many applications. A camera that could ``see around corners'' could help improve navigation and mapping capabilities of autonomous vehicles or make search and rescue missions more effective. Time-resolved single-photon imaging systems have recently been demonstrated to record optical information of a scene that can lead to an estimation of the shape and reflectance of objects hidden from the line of sight of a camera. However, existing non-line-of-sight (NLOS) reconstruction algorithms have been constrained in the types of light transport effects they model for the hidden scene parts. We introduce a factored NLOS light transport representation that accounts for partial occlusions and surface normals. Based on this model, we develop a factorization approach for inverse time-resolved light transport and demonstrate high-fidelity NLOS reconstructions for challenging scenes both in simulation and with an experimental NLOS imaging system. 
\end{abstract}

\begin{teaserfigure}  
  \includegraphics[width=\linewidth]{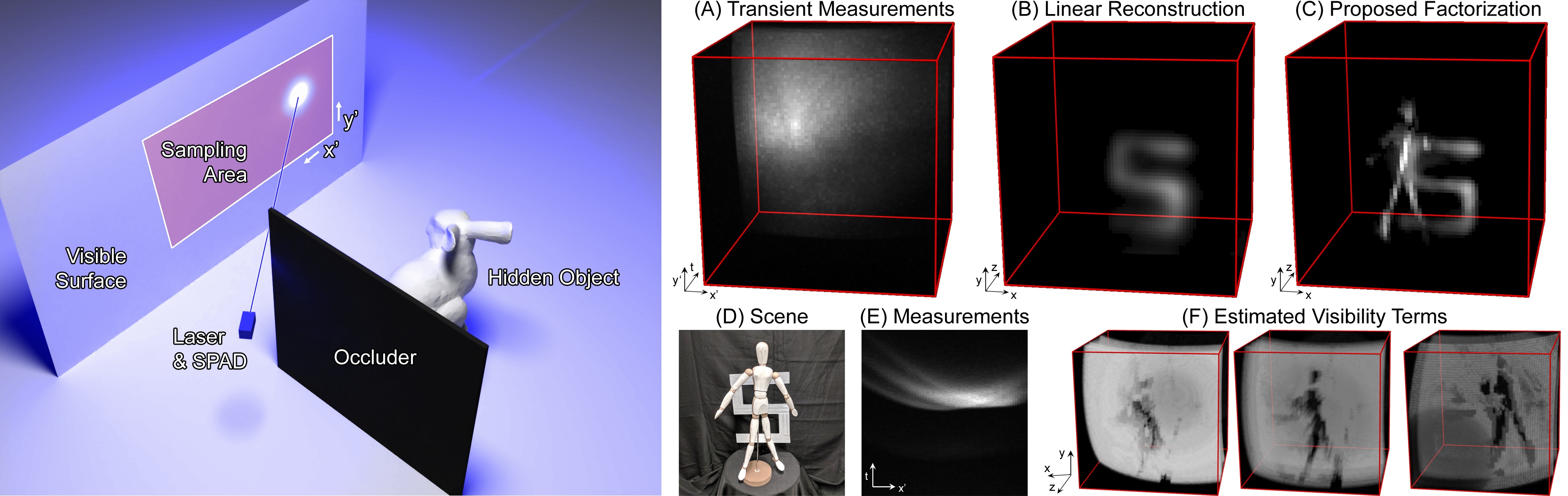}
  \caption{Non-line-of-sight (NLOS) imaging aims at recovering the shape and albedo of objects hidden from a camera or a light source. Using ultra-fast pulsed illumination and single photon detectors, the light transport in the scene is sampled for visible objects (left). The global illumination components of these time-resolved measurements (A,E) contain sufficient information to estimate the shape of hidden objects (B,C). Using a novel formulation for NLOS light transport that models partial occlusions of hidden objects (D) via visibility terms (F), we demonstrate higher-fidelity reconstructions (C) than previous approaches to NLOS imaging (B).}
  \label{fig:teaser}
\end{teaserfigure}

%
%



\maketitle
\thispagestyle{empty}


\newcommand{\gw}[1]{\textcolor{red}{(gw: #1)}}
\newcommand{\mo}[1]{\textcolor{red}{(mo: #1)}}
\newcommand{\re}[1]{\textcolor{red}{#1}} 

\newcommand{\vectorsym}[1]{\mathbf{#1}}
\newcommand{\matrixsym}[1]{\mathbf{#1}}
\newcommand{\rvarsym}[1]{#1}

\newcommand{\ft}{\mathcal{F}_t}

\newcommand{\albedoisotropic}{\rho}
\newcommand{\albedo}{\rho}
\newcommand{\albedod}{\boldsymbol{\rho}}

\newcommand{\ttime}{\tau}               	
\newcommand{\ttimed}{\boldsymbol{\tau}}  	
\newcommand{\npixels}{N}               		

\newcommand{\cpy}{\boldsymbol{\mathcal{I}}}

\newcommand{\msys}{\mathbf{A}}
\newcommand{\mvis}{\mathbf{V}}
\newcommand{\mtrans}{\mathbf{T}}
\newcommand{\mnorm}{\mathbf{N}}
\newcommand{\normals}{\mathbf{n}}

\newcommand{\veryshortrightarrow}[1][3pt]{\mathrel{%
   \hbox{\rule[\dimexpr\fontdimen22\textfont2-.2pt\relax]{#1}{.4pt}}%
   \mkern-4mu\hbox{\usefont{U}{lasy}{m}{n}\symbol{41}}}}

\newcommand{\admmparameter}{\varrho}

\newcommand{\convop}{*}
\newcommand{\timevar}{t}
\newcommand{\traveltimevar}{\tau}
\newcommand{\signalfun}[1]{i({#1})}
\newcommand{\pulsefun}[1]{g({#1})}

\newcommand{\ind}{\mathcal{I}}

\newcommand{\flux}{r}
\newcommand{\numphotons}{\lambda}

\newcommand{\pdp}{\eta}
\newcommand{\darkcount}{d}
\newcommand{\pulse}{g}
\newcommand{\jitter}{f}
\newcommand{\histogram}{h}
\newcommand{\pois}{\mathcal{P}}

\newcommand{\imageformationop}{J}

\newcommand{\meas}{\vectorsym{\histogram}}
\newcommand{\unknown}{\boldsymbol{\rho}}

\newcommand{\conjugateop}[1]{{#1}^*}
\newcommand{\transposeop}[1]{{#1}^T}
\newcommand{\avgop}[1]{\langle {#1} \rangle}
\newcommand{\innerop}[1]{\langle {#1} \rangle}
\newcommand{\absop}[1]{\left| {#1} \right|}

\newcommand{\expectation}[1]{\mathbf{E} \left[ {#1} \right]}

\newcommand{\Bexp}[1]{\exp\left( {#1} \right)}

\newcommand{\field}[1]{E_{#1}}
\newcommand{\norm}[1]{\left\lVert#1\right\rVert}

\newcommand{\numlights}{N}

\newcommand{\photovector}{\vectorsym{p}}

\newcommand{\densitymatrix}{\matrixsym{T}}
\newcommand{\lightmatrix}{\matrixsym{L}}
\newcommand{\lightfreq}{\omega}

\newcommand{\oforder}{\sim}

\newcommand{\warning}[1]{\textcolor{blue}{#1}}

\newcommand{\argmin}[1]{\stackrel[#1]{}{\textrm{ arg min}}}
\newcommand{\argmax}[1]{\stackrel[#1]{}{\textrm{ arg max}}}
\newcommand{\minimize}[1]{\stackrel[#1]{}{\textrm{minimize}}}

\newcommand{\etal}{et al.}


\definecolor{Gray}{rgb}{0.5,0.5,0.5}
\definecolor{darkblue}{rgb}{0,0,0.7}
\definecolor{orange}{rgb}{1,.5,0} 
\definecolor{red}{rgb}{1,0,0} 

\newcommand{\heading}[1]{\noindent\textbf{#1}}
\newcommand{\note}[1]{{\em{\textcolor{orange}{#1}}}}
\newcommand{\todo}[1]{{\textcolor{darkblue}{TODO: #1}}}
\newcommand{\comments}[1]{\small{\emph{\textcolor{Gray}{#1}}}}
\newcommand{\place}[1]{ \begin{itemize}\item\textcolor{darkblue}{#1}\end{itemize}}
\newcommand{\de}{\mathrm{d}}

\newcommand{\Radiance}{\mathit{L}}             

\newcommand{\Radiosity}{B}             

\newcommand{\Visible}{V}             
\newcommand{\Refl}{\mathit{\rho}}             
\newcommand{\Geom}{\mathit{F}}             
\newcommand{\geom}{g}             

\newcommand{\volume}{V}
\newcommand{\normal}{\Vect{n}}
\newcommand{\laserpos}{\Vect{l}}               
\newcommand{\wallpos}{\Vect{w}}               
\newcommand{\voxelpos}{\Vect{x}}      
\newcommand{\pixelpos}{\Vect{c}}               
\newcommand{\pathlength}{p}
\newcommand{\posvariable}{\Vect{y}}
\newcommand{\transimg}{\Vect{i}}

\newcommand{\timecoord}{t}

\newcommand{\Mat}[1]    {{\matrixsym{#1}}} 
\newcommand{\Pen}      		{F} 
\newcommand{\cardset}     {\mathcal{C}}
\newcommand{\Dat}      		{G} 

\section{Introduction}
\label{sec:intro}
The capacity of imaging systems must continue to expand to keep pace with rapidly emerging technologies. Autonomous vehicles, for example, would greatly benefit from improved vision in fog, snow, and other scattering media or from being able to see around corners to detect what lies beyond the next bend or another car. Sensing technology offering such non-line-of-sight (NLOS) capabilities could help make self-driving cars safer and unlock unprecedented potential for other machine vision systems. This is a broad vision towards which this paper makes a significant step.

Two challenges make NLOS imaging with time-resolved detectors difficult. First, the low signal of multiply scattered light places extreme requirements on photon sensitivity of the detectors. Second, inverse methods that aim at estimating shape, albedo, and other properties of a hidden scene need to model the transient light transport appropriately and devise means to robustly invert it. To address the former issue, we follow recent work proposing acquisition setups with a pulsed picosecond laser and single-photon avalanche diodes (SPADs) for NLOS imaging and tracking~\cite{buttafava2015,gariepy2016,Chan:17,OToole:2018}. SPADs are detectors that digitize the time of arrival of individual photons with a precision in the order of tens of picoseconds, thus resolving light paths with approx. centimeter resolution. At the core of our paper is a factored representation of transient light transport that models partial occlusions and surface normals in the hidden scene. These effects have been mostly ignored by other NLOS imaging approaches~\cite{velten2012,Gupta:12,Wu:2012:frequency,Heide:2014:diffusemirrors,buttafava2015,OToole:2018}. Moreover, we derive a robust multi-convex reconstruction algorithm that takes a measured transient image as input and factors it into the proposed representation: a volume of hidden albedos and surface normals along with visibility terms that model partial occlusion in the hidden scene (see Fig.~\ref{fig:teaser}).

With the presented work, we take first steps towards making NLOS imaging robust and practical for real-world applications. Specifically, we make the following contributions:
\begin{itemize}
	\item We introduce a factored nonlinear image formation model for non-line-of-sight imaging that accounts for partial occlusions and surface normals in the hidden scene.
	\item We derive a multi-convex solver for inverse transient light transport and show that it achieves significantly higher reconstruction quality than conventional NLOS imaging methods.
	\item We implement an experimental NLOS acquisition setup using a single-photon avalanche diode and a picosecond laser. 
	\item We validate the proposed reconstruction algorithms in simulation and with example scenes captured with the prototype.
\end{itemize}

\paragraph{Overview of Limitations}

Although the proposed inverse method improves reconstruction quality for many types of hidden scenes, it is also computationally more expensive than other methods. Specifically, the memory requirements of the proposed factorization method are two orders of magnitude higher than for matrix-free implementations of simpler inverse methods. Similar to other non-line-of-sight methods, we make the assumption that the measured transient light transport contains only first-order and third-order bounces. The first-order bounces correspond to direct illumination that is reflected off a visible wall back to the detector; these contributions can be removed from the measurements and also used to estimate surface normals and albedos of the visible wall. The third-order bounces contain indirect illumination that bounced precisely three times before reaching the detector: off the visible wall, then off a hidden object, then off the visible wall again.

\section{Related Work}
\label{sec:related}
\paragraph{Non-line-of-sight Imaging}

Kirmani~\etal~\shortcite{Kirmani:2009} first introduced the idea of ``looking around corners'' by analyzing the feasibility of reconstructing hidden objects from time-resolved light transport. This concept was demonstrated in practice by Velten~\etal~\shortcite{velten2012} with a system capable of resolving the shape and reflectance of a hidden object. Velten's hardware setup included a streak camera and a femtosecond laser, which together account for a cost of several hundred thousand dollars. The streak camera provides a theoretical precision of up to 2~ps, which corresponds to a travel distance of 0.6~mm. Correlation-based time-of-flight sensors have also been demonstrated as a low-cost alternative for non-line-of-sight imaging~\cite{kadambi2013,Heide:2014:diffusemirrors}. While these systems are about three orders of magnitude less expensive than Velten's system, they also only offer a very limited temporal resolution, thus limiting the quality of reconstructed NLOS scenes. Recently, single photon avalanche diodes (SPADs) have been proposed for NLOS imaging~\cite{buttafava2015,OToole:2018} as a readily-available hardware platform that offers a good balance between cost and precision.
 
NLOS imaging requires a model for the light transport of hidden scene parts as well as a large-scale reconstruction framework. Existing proposals for NLOS imaging~\cite{Kirmani:2009,velten2012,Gupta:12,Wu:2012:frequency,Heide:2014:diffusemirrors,buttafava2015,OToole:2018} use an image formation model that makes the following assumptions: (1) light bounces at most three times within the scene; (2) the scene contains no occlusions; (3) light scatters isotropically (i.e., surface normals are ignored). Under these assumptions, the reconstruction becomes a linear inverse problem. Velten et al.~\shortcite{velten2012}, Gupta et al.~\shortcite{Gupta:12}, Buttafava et al.~\shortcite{buttafava2015}, and Jarobo et al.~\shortcite{Arellano:17} solved this system using variants of the backprojection algorithm. Due to the fact that this tends to emphasize low frequencies and does not actually solve the inverse problem, the reconstruction quality offered by backprojection-type inverse methods tends to be low. Using the same assumptions, Gupta et al.~\shortcite{Gupta:12}, Wu et al.~\shortcite{Wu:2012:frequency}, Heide et al.~\shortcite{Heide:2014:diffusemirrors}, and~\cite{OToole:2018} proposed to solve the inverse problem via large-scale iterative optimization. While this approach is more accurate than backprojection, the underlying light transport model ignores partial occlusions and surface normals, which we show to be crucial for accurate scene reconstruction. Finally, Tsai et al.~\shortcite{Tsai:2017} recently proposed a space carving algorithm for estimating the convex hull of hidden objects; a full 3D volume of the hidden scene cannot be recovered with this approach. 

At the core of this paper is a novel image formation model that models NLOS light transport more accurately than existing methods by accounting for partial occlusions and surface normals in the hidden scene; we derive inverse methods tailored to this model.

\paragraph{Single Photon Avalanche Diodes} SPADs are reverse-biased photodiodes that are operated well above their breakdown voltage (see e.g.~\cite{burri2016}). Every photon incident on a SPAD has some probability of triggering an electron avalanche which is time-stamped. This time-stamping mechanism usually provides an accuracy of tens to hundreds of picoseconds. SPADs and also avalanche photodiodes (APDs) are commonly used for a wide range of applications in optical telecommunication, fluorescence lifetime imaging, and remote sensing systems (e.g., LIDAR). Often, these imaging modes are referred to as time-correlated single photon counting~\cite{OConnor:2012}. 

Recently, SPADs were applied to range imaging~\cite{kirmani2014,Shin:2016}, transient imaging~\cite{gariepy2015,OToole:2017:SPAD} as well as tracking~\cite{gariepy2016,Chan:17} and imaging~\cite{buttafava2015,OToole:2018} non-line-of-sight objects. The works by Buttafava et al.~\shortcite{buttafava2015} and~\cite{OToole:2018} are closest to ours, but their reconstruction algorithms ignore hidden surface normals and occlusions. While the approach proposed by~\cite{OToole:2018} is computationally efficient by modeling NLOS light transport as a shift-invariant convolution, normals and the visibility term modeling partial occlusions are not suitable for this representation because these effects create spatial variation in the image formation.

\paragraph{Imaging Through and Around Stuff}

Other forms of non-line-of-sight imaging have also been demonstrated that do not rely on time-resolved imaging. For example, Sen et al.~\shortcite{Sen:2005} proposed a projector-camera system where the viewpoints of camera and projector could be interchanged. This approach allows the scene to be hidden from the camera's perspective, but it must be visible from the projector's perspective. In time-resolved NLOS imaging, the scene is typically not directly observed from either detector or light source. Snapshot NLOS imaging was demonstrated by exploiting correlations that exist in coherent laser speckle~\cite{katz2014}, though this has so far been demonstrated at microscopic scales. Radio and terahertz frequencies were shown to be able to imaging through objects due to the physical properties of these parts of the electromagnetic spectrum~\cite{Adib2015,RedoSanchez2016}. Recovering and tracking hidden objects was also shown to be possible with intensity measurements of conventional cameras to a limited extent~\cite{Klein:2016,Bouman:2017}.

In concurrent work, Thrampoulidis et al.~\shortcite{Thrampoulidis:2017} developed an alternative model that also includes partial occlusions for NLOS imaging. As opposed to our model, this work assumes that the shape of the NLOS scene, including occluders, is known and only surface albedos need to be recovered. We make no such assumptions; we model and recover unknown shape, albedo, visibility and also surface normals.

\section{Forward and Inverse Light Transport}
\label{sec:theory}
\begin{figure}
	\centering
		\includegraphics[width=0.9\columnwidth]{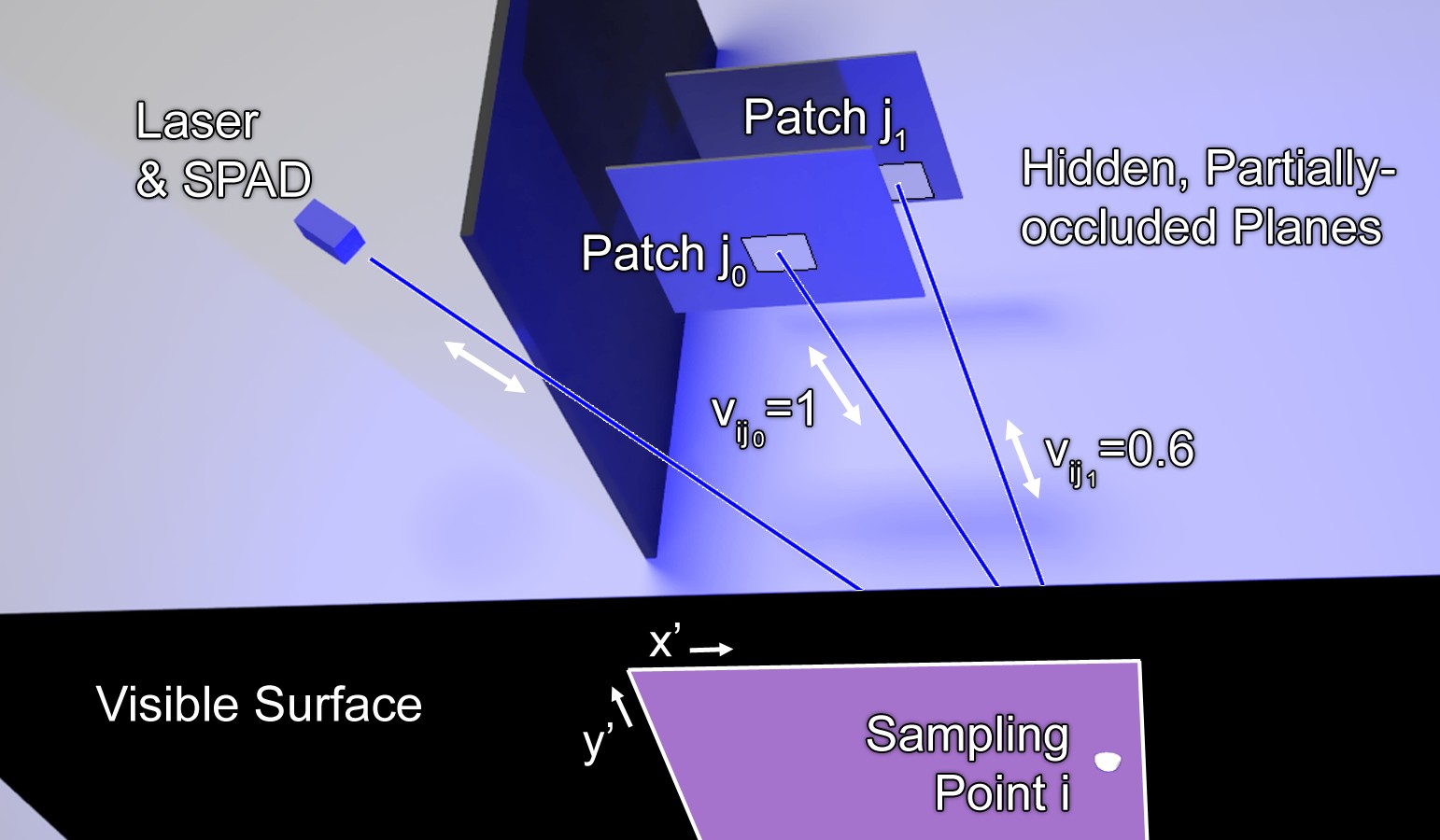}
		\caption{NLOS scene with partial occlusions. The detector and laser sample the visible wall at point $i$ to record the direct and indirect light transport. The indirect components include contributions from hidden objects, such as two patches $j_0$ and $j_1$. Whereas the optical path between $i$ and $j_0$ is unoccluded, the path between $i$ and $j_1$ is partly obscured by another hidden surface. These occlusions are modeled by a visibility term $v$.}
		\label{fig:setup}
\end{figure}

The non-line-of-sight imaging problem involves estimating 3D shape and albedo of objects outside the line of sight of a detector from third-order bounces of time-resolved global light transport. Specifically, a short light pulse is focused on a visible part of the scene, for example a wall, the light scatters off that surface, reaches a hidden object which scatters some of the light back to the visible surface, where it can be recorded with a time-resolved detector. While several different acquisition setups have been proposed, each warranting a slightly different image formation model, we follow O'Toole et al.~\shortcite{OToole:2018} and model a confocal system, where a single time-resolved detector is co-axially aligned with a pulsed light source to sample positions $x', y'$ on a visible diffuse wall (see Figs.~\ref{fig:teaser},~\ref{fig:setup}). 

\subsection{Confocal NLOS Image Formation Model}

A time-resolved detector measures the incident photon flux as a function of time relative to an emitted light pulse. Such a detector can be used to record the temporal impulse response of a scene, including direct and global illumination, at sampling positions $x',y'$ on a visible surface, resulting in a 3D space-time volume that is known as a \emph{transient image} $\ttime$. 

The direct illumination, i.e., light emitted by the source and scattered back to the detector from an object, contains all information necessary to recover the shape and reflectance of visible scene parts. This is commonly done for 3D imaging or light detection and ranging (LIDAR)~\cite{Schwarz:10,McCarthy:13,kirmani2014,Shin:2016}. In the following image formation model, the direct light is not considered because it can be removed from measurements acquired in practice; only the global illumination contains useful information for non-line-of-sight imaging. 

The image formation model can be formulated as
\begin{align}\label{eq:math:cnlos}
	\ttime \left( x', y', t \right) & = \int \!\!\!\!  \int \!\!\!\! \int_{\Omega} \,\, \frac{1}{r^4} \,\, \albedoisotropic \left( x,y,z \right) \,\, \\
	& \delta \left( 2\sqrt{\left( x' - x \right)^2 + \left( y' - y \right)^2 + z^2} - tc \right)  \,\, dx \, dy \, dz, \nonumber	
\end{align}
where the Dirac delta function $\delta\left( \cdot \right)$ relates the time of flight $t$ to the distance function $r = \sqrt{ \left( x' - x \right)^2 + \left( y' - y \right)^2 + z^2 } = tc / 2$. Here, $c$ is the speed of light. This image formation model makes several assumptions on the light transport in the hidden scene: light scatters only once (i.e., back to the visible scene parts), light scatters isotropically (i.e., surface normals are ignored), and no occlusions occur between different scene parts outside the line of sight. 

We lift two of these assumptions by augmenting Equation~\ref{eq:math:cnlos} by a visibility term $v$ as well as surface normals $n$:
\begin{align}
	\ttime \left( x', y', t \right)  & = \left( \frac{2}{tc} \right)^4 \!\! \int \!\!\!\! \int \!\!\!\! \int_{\Omega} \,\, v \left( x', y', x, y, z \right) \albedo \left( x,y,z\right) \,\, \left( \omega \cdot n \left( x, y, z \right) \right) \nonumber	\\
	& \delta \left( 2\sqrt{\left( x' - x \right)^2 + \left( y' - y \right)^2 + z^2} - tc \right)  \,\, dx \, dy \, dz, 
	\label{eq:math:cnlosvis}
\end{align}
The distance falloff $1/r^4$ (see Eq.~\ref{eq:math:cnlos}) is replaced by the factor $( 2 / tc )^4$, which can be pulled out of the integral because it is space-invariant for confocal scanning setups. The term $v \left( x', y', x, y, z \right) \in [0,1]$ models the visibility of a hidden surface patch at location $x,y,z$ observed from the position $x',y'$ on the visible wall. For partial occlusion in the hidden scene, such a patch may be visible from one sampling point on the wall but it may be occluded by another hidden object from the perspective of another sampling point (see Fig.~\ref{fig:setup}).

In a confocal scanning setup, the bidirectional reflectance distribution function (BRDF) $f_r$ of the hidden scene is only sampled at a subset of all combinations of incident and outgoing light direction: $\omega_i = \omega_o = \omega$, where $\omega$ is the normalized direction from a location $x,y,z$ to some sampling point $x',y'$. Thus, it may be infeasible to recover arbitrary BRDFs from confocal measurements. However, when the BRDF can be modeled as a spatially-varying but directionally constant albedo $\albedo \left( x,y,z\right) = f_r \left( x,y,z, \omega, \omega \right)$, which is the case for diffuse and also retroreflective materials, this albedo can indeed be estimated as shown in previous work. Retroreflective BRDFs have the benefit of increasing the amount of light reflected from hidden scene parts back to wall by two orders of magnitude, such that the $1/r^4$ distance falloff becomes $1/r^2$ (cf. Eq.~\ref{eq:math:cnlos}). 

\subsection{Factored Image Formation}

\begin{figure}[t]
	\centering
		\includegraphics[width=0.8\columnwidth]{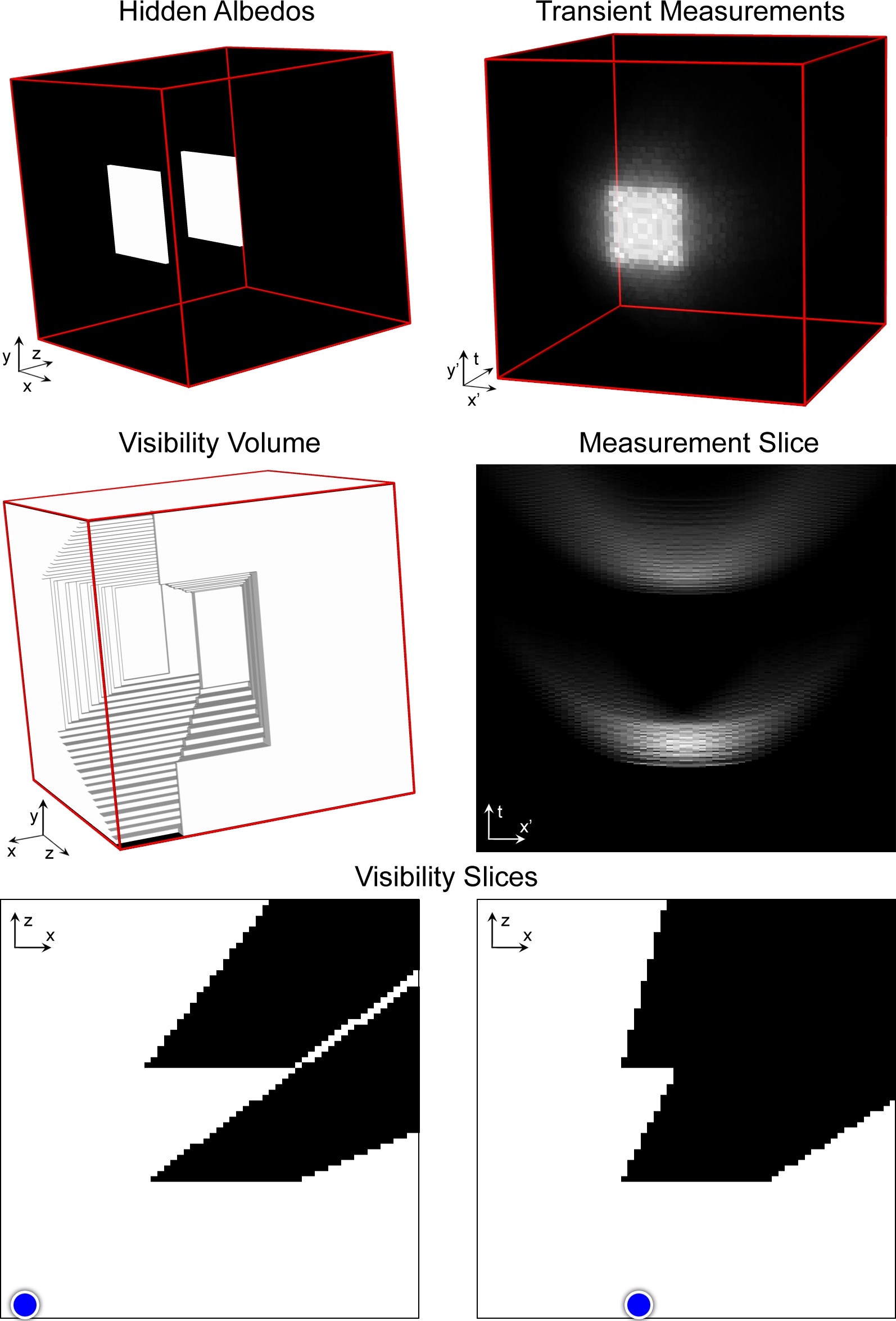}
		\caption{Illustration of several light transport terms for a simple hidden scene containing two partially-occluding white planes (top left). The transient image of measurements is shown (top right) along with a single xt-slice (center right). A 3D rendering of the ground truth visibility term for one sampling location (center left) and two xz-slices (bottom) for different sampling positions (blue circles) make it intuitive to understand what the visibility terms are.}
		\label{fig:factors}
\end{figure}

We discretize Equation~\ref{eq:math:cnlosvis} by representing the hidden volume as $N \times N \times N$ voxels. Each voxel $j = 1 \ldots N^3$ contains an albedo $\albedod_j$ and a surface normal $\normals_j$. The discrete transient image is sampled at $N \times N$ locations that coincide with the voxel centers on the visible wall. For notational convenience, we model the transient image with $N$ temporal bins at each spatial location. The image formation model becomes 
\begin{equation}
	\ttimed = \msys \albedod = \left( \mtrans \circ \cpy \left( \mnorm \circ \mvis \right) \right) \albedod,
	\label{eq:discrete_image_formation}
\end{equation}
where $\ttimed \in \mathbb{R}_+^{N^3}$ is the vectorized transient image and $\albedod \in \mathbb{R}_+^{N^3}$ is the vectorized volume of nonnegative hidden albedos. The system matrix $\msys \in \mathbb{R}_+^{N^3 \times N^3}$ combines all others terms of the transient light transport (cf.~Eq.~\ref{eq:math:cnlosvis}). This matrix representation has also been used in most previous approaches to NLOS imaging. We propose to factor the transient light transport matrix $\msys$ into several terms, each modeling different aspects of light transport, as discussed in the following and illustrated in Figure~\ref{fig:factors}. 

\paragraph{Visibility} The visibility term $\mvis \in \mathbb{R}_+^{N^2 \times N^3}$ is time-invariant and models how much of the light reflected by voxel $j$ reaches measurement location $i = 1 \ldots N^2$ on the visible wall. As shown in Figure~\ref{fig:setup}, when the path between $j$ and $i$ is unoccluded: $\mvis_{ij}=1$. When another surface occludes the path between $j$ and $i$: $\mvis_{ij}=0$. We allow for a continuous range of values, i.e. $0 \leq \mvis_{ij} \leq 1$, to model partial occlusion along a light path. 

\paragraph{Normals} The matrix $\mnorm \in \mathbb{R}^{N^2 \times N^3}$ is also time-invariant and models the factor $\omega \cdot \normals$, such that $\mnorm_{ij} = \omega_{j \rightarrow i} \cdot \normals_j$ where $\omega_{j \rightarrow i}$ is the normalized direction pointing from voxel $j$ to the visible wall location $i$. For the purpose of this paper, we parameterize the surface normals in spherical coordinates. That is, the normal $\normals_j$ at voxel $j$ is represented using two scalars $\normals_j^{u},\normals_j^{v}$, such that $\normals_j = [ cos(\normals_j^{u}) \, sin(\normals_j^{v}),\,\,$ $sin(\normals_j^{u}) \, cos(\normals_j^{v}),\,\,$ $cos(\normals_j^{v}) ]^T$. This representation enforces unit length on all surface normals and only requires two, instead of three, parameters to be estimated per hidden normal.

\paragraph{Copy Matrix} To account for the fact that neither $\mvis$ nor $\mnorm$ are time-dependent, but the transport matrix $\mtrans$ is, the matrix $\cpy \in \mathbb{R}_+^{N^3 \times N^2}$ simply copies the time-independent quantities of the hidden volume projected on the sampling locations to all time bins of the transient image. 

\paragraph{Transport Matrix} The matrix $\mtrans \in \mathbb{R}_+^{N^3 \times N^3}$ models the time-dependent aspects of the hidden light transport. Specifically, the $j^{th}$ column of $\mtrans$ is the surface of the hypercone $\left( x' - x_j \right)^2 + \left( y' - y_j \right)^2 + z_j^2 = (tc/2)^2$ modeling time-resolved propagation in free space from voxel $j$ to the entire transient image. The super-position principle holds, such that the hypercones for each voxel contribute to the transient image in an additive way. \cite{OToole:2018} recently showed how to model NLOS light transport without occlusions or normals as a shift-invariant convolution, which leads to a closed-form solution of the NLOS problem. Unfortunately, this model is not applicable in our scenario because accounting for the visibility terms and surface normals make this a spatially-varying image formation model.

\subsection{Inverse NLOS Light Transport}

Several inverse methods for the non-line-of-sight imaging problem have been proposed. We briefly review backprojection-type methods and linear inverse methods before introducing a factorization approach that allows us to recover the unknown visibility terms and surface normals along with the hidden albedos. 

\subsubsection{Backprojection}

The NLOS problem can be reduced to a linear one under certain assumptions. First, the visibility term is ignored (i.e., $V_{ij}=1, \forall i,j$) and, second, the surface normals are fixed. For the latter, one could either make the assumption that the hidden scene is comprised of isotropic scatterers (i.e., $N_{ij}=1, \forall i,j$) or assume that the hidden normals are known, but that is typically not the case. Previous work on NLOS imaging has shown that these assumptions lead to the linear image formation model $\ttimed = \msys \albedod$, with $\msys = \mtrans$ (cf.~Eq.~\ref{eq:discrete_image_formation}). 

Filtered and unfiltered backprojection methods are standard algorithms for solving many linear inverse problems, particularly in computed tomography~\cite{Kak:1988}. The beauty of backprojection methods is their simplicity, i.e., both compute time ($O\left( N^5 \right)$) and memory requirements (matrix free implementation $O\left( N^3 \right)$; with sparse matrix $O\left( N^5 \right)$) are tractable even for large-scale inverse problems. Velten et al.~\shortcite{velten2012}, Gupta et al.~\shortcite{Gupta:12}, Buttafava et al.~\shortcite{buttafava2015}, and Arellano et al.~\shortcite{Arellano:17} all employ a variant of backprojection by multiplying the measured transient image by the transpose of the system matrix, i.e., $\albedod \approx \msys^T \ttimed$, and then optionally applying a sharpening filter, such as a Laplacian, and a thresholding operator~\cite{velten2012}. 

Unfortunately, filtered backprojection only solves the linear problem correctly when measurements over the full sphere are available. The acquisition setups of NLOS imaging discussed in the literature resemble that of a limited-baseline tomography problem, for which backprojection only gives a rough estimate of the latent variable, but it does not solve the actual inverse problem.

\subsubsection{Linear Inverse Light Transport}

Several other NLOS reconstruction algorithms~\cite{Gupta:12,Wu:2012:frequency,Heide:2014:diffusemirrors} solve the system of linear equations directly, but they make the same assumptions on visibility and normals as the backprojection algorithm. The inverse problem of recovering hidden albedos can be expressed as 
\begin{equation}
	\minimize{\albedod} \left\| \ttimed- \msys \albedod \right\|_2^2 + \Gamma \left( \albedod \right), \,\,\, \textrm{s.t.} \,\,0 \leq \albedod
	\label{eq:objective_noocclusions}
\end{equation}
Although the nonnegativity constraints were not directly enforced by all previous proposals, including it in the reconstruction can improve the estimated solution. An additional prior on the albedos $\Gamma \left( \albedod \right)$ can help further improve the estimated albedos. For example, Heide et al.~\shortcite{Heide:2014:diffusemirrors} used a combination of sparseness and sparse gradients (i.e., total variation). The runtime and memory requirements for an iterative solver are in the same order as those of the backprojection method per iteration.

\subsubsection{Factorized Light Transport}

Assuming that neither the hidden albedos $\albedod$, surface normals $\normals$, or visibility terms $\mvis$ are known, inverting Equation~\ref{eq:discrete_image_formation} becomes a nonlinear inverse problem with the cost function
\begin{equation}
\begin{aligned}
	\minimize{\albedod, \normals, \mvis} & \left\| \ttimed - \left( \mtrans \circ \cpy  \left( \mnorm \circ \mvis \right) \right) \albedod \right\|_2^2 + \Gamma \left( \albedod \right) . \\
	& \textrm{s.t.} \,\, 0 \leq \mvis \leq 1, \quad 0 \leq \albedod
\end{aligned}
	\label{eq:objective_occlusions}
\end{equation}
%

An important insight for solving Equation~\ref{eq:objective_occlusions} efficiently is that although the cost function is nonlinear, it is tri-convex when the prior $\Gamma$ is convex. As is standard practice for multi-convex problems, we use an alternating least-squares (ALS) approach. To this end, Equation~\ref{eq:objective_occlusions} is solved in an alternating manner by fixing two of the unknown terms and optimizing for the third. Each of these subproblems is convex; the method is outlined in Algorithm~\ref{alg:als}. 

\begin{algorithm}[h!]
\caption{Triconvex Factorization via Alternating Least Squares}
\label{alg:als}
\begin{algorithmic}[1]
\State $\mvis^{(0)} = \mathbf{1}, \,\, \mnorm^{(0)} = rand(), \,\, \albedod^{(0)} = rand()$
\State $\textrm{\bfseries for} \,\, k=1 \,\, \textrm{\bfseries to } \, K$ 
\State $\,\,\, \albedod^{(k)} \leftarrow \argmin{0 \leq \albedod } \left\| \ttimed - \left( \mtrans \circ \cpy  \left( \mnorm^{(k-1)} \circ \mvis^{(k-1)} \right) \right) \albedod \right\|_2^2  + \Gamma_{\albedod} \left( \albedod \right)$
\State $\,\,\, \mathbf{V}^{(k)} \leftarrow \argmin{0 \leq \mathbf{V} \leq 1} \left\| \ttimed - \left( \mtrans \circ \cpy \left( \mnorm^{(k-1)} \circ \mvis \right) \right) \boldsymbol{\rho}^{(k)}  \right\|_2^2$
\State $\,\,\, \mathbf{\normals}^{(k)} \!\! \leftarrow \argmin{\normals} \left\| \ttimed - \left( \mtrans \circ \cpy \left( \mnorm \circ \mvis^{(k)} \right) \right) \albedod^{(k)}  \right\|_2^2 \,\, + \Gamma_{n} \left( \normals \right)$
\State $\textrm{\bfseries end} \,\, \textrm{\bfseries for}$	
\end{algorithmic}
\end{algorithm}

\paragraph{Updating $\albedod$ (Alg.~\ref{alg:als}, line~3)} In this subproblem, the system matrix $\msys = \mtrans \circ \cpy  \left( \mnorm^{(k-1)} \circ \mvis^{(k-1)} \right) $ is fixed for a given iteration $k$. The resulting inverse problem is similar to that of Equation~\ref{eq:objective_noocclusions}; we use the alternating direction method of multipliers (ADMM)~\cite{Boyd:2011} to solve it. 

\paragraph{Updating $\mvis$ (Alg.~\ref{alg:als}, line~4)}

This subproblem is also convex because we could construct a system matrix that absorbs $\mtrans, \mnorm, \cpy$ and $\albedod$ and write the image formation as a matrix-vector multiplication. Unfortunately, the size of this problem is very large -- the system matrix would have $N^3 \times N^5$ non-zero elements. Thus, we solve this subproblem using a projected gradient algorithm that minimizes the objective function $J = \left\| \ttimed - \left( \mtrans \circ \cpy \left( \mnorm \circ \mvis \right) \right) \albedod \right\|_2^2$ while enforcing $0 \leq \mvis \leq 1$. For this algorithm, we simply take a step into the direction of the negative gradient of $J$ and clamp the result to the feasible range, i.e., between 0 and 1, using the projection operator $\Pi$:
\begin{align} \label{eq:objective_occlusions_gradient}
J & \leftarrow \Pi \left( J - \alpha \nabla_{v} J \right), \,\,\, \textrm{with} \\
\nabla_{v} J & = \,\, -2 \mnorm \circ \cpy^T \left( \left( \left( \ttimed - \left( \mtrans \circ \cpy \left( \mnorm \circ \mvis \right) \right) \albedod \right)  \albedod^T \right) \circ \mtrans\right) \nonumber
\end{align}
Here, $\alpha$ is the step length. In practice, this intuitive method can be improved using an adaptive step length which changes per iteration. We we derive this in detail in the Supplemental Material.

\paragraph{Updating $\normals$ (Alg.~\ref{alg:als}, line~5)}

Using the spherical coordinate representation of the normals, this subproblem can be solved with an unconstrained nonlinear solver to minimize the objective function $J$ with respect to $\normals^u,\normals^v$. We chose the L-BFGS algorithm for this task and, using the chain rule, derive the gradient of the objective as
\begin{align}
\nabla_{\normals^{(u,\!v)}} J & = \nabla_{\mnorm} J \cdot \nabla_{\normals^{(u,\!v)}} \mnorm, \quad \text{with}  \\
\nabla_{\mnorm} J & = -2 \mvis \circ \cpy^T \left( \left( \left( \ttimed - \left( \mtrans \circ \cpy \left( \mnorm \circ \mvis \right) \right) \albedod \right) \albedod^T \right) \circ \mtrans\right), \nonumber \\
\nabla_{\!\!\normals^{(u,\!v)}} \! \mnorm_{ij} \!\! & = \!\! 
\begin{bmatrix} 
-\omega_{j \veryshortrightarrow \! i}^{x} \sin( \normals_j^{u} ) \sin ( \normals_j^{v} ) \!+\! \omega_{j \veryshortrightarrow \! i}^{y} \cos ( \normals_j^{u} ) \cos ( \normals_j^{v} )   \nonumber \\
\!\omega_{j \veryshortrightarrow \! i}^{x} \cos ( \normals_j^{u} ) \cos ( \normals_j^{v} ) \!- \!\omega_{j \veryshortrightarrow \! i}^{y} \sin ( \normals_j^{u} ) \sin ( \normals_j^{v} ) \!- \!\omega_{j \veryshortrightarrow \! i}^{z} \sin ( \normals_j^{v} ) \!
\end{bmatrix}^T \nonumber
\end{align}

%

\section{Evaluation}
\label{sec:simulations}
\begin{figure*}[t!]
	\centering
		\includegraphics[width=0.95\textwidth]{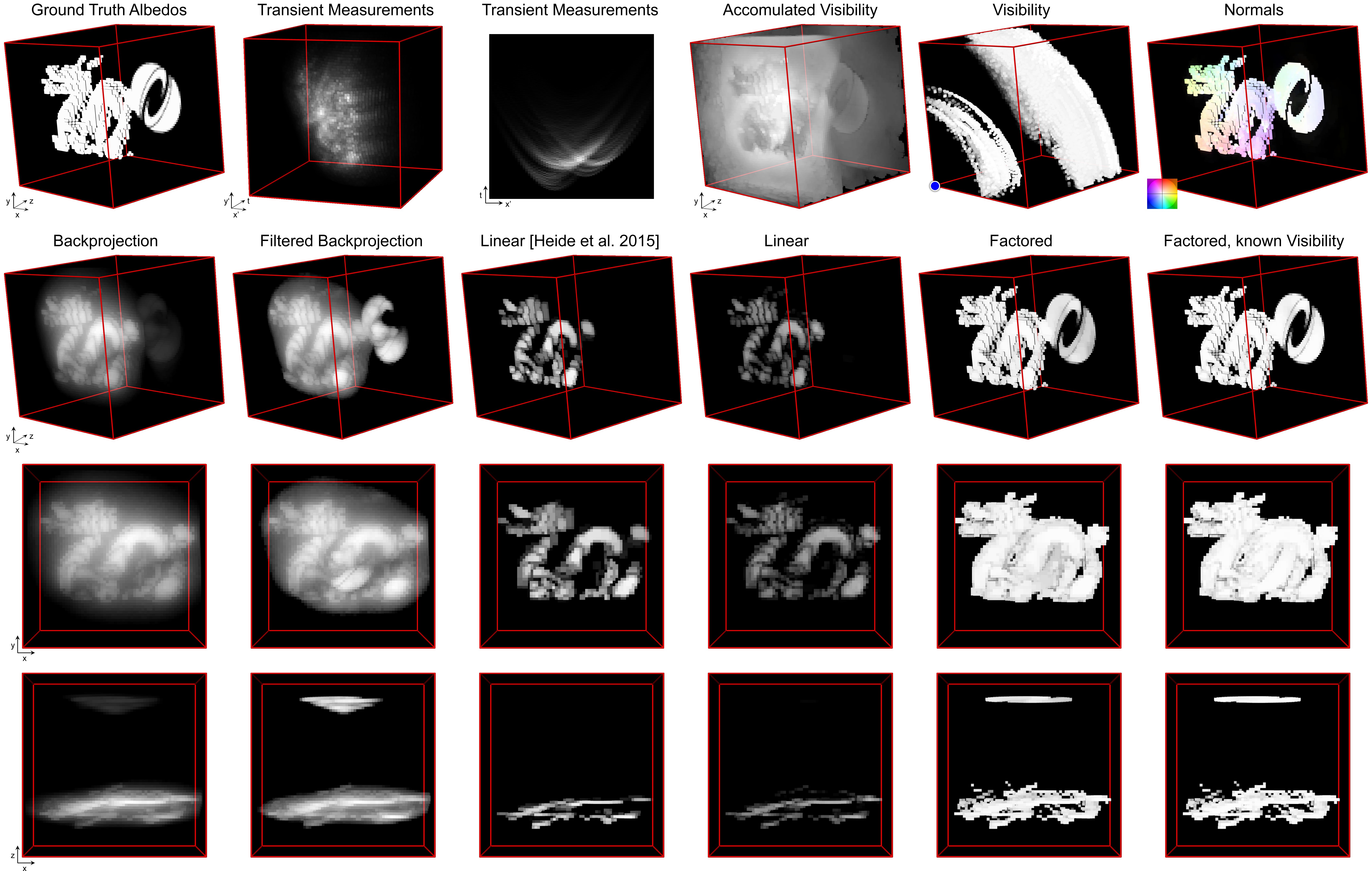}
		\caption{Evaluation of reconstruction algorithms using the ``Dragon \& Logo'' scene. The hidden scene contains two objects that partially occlude each other. Time-resolved measurements are shown along with reconstructions obtained with several algorithms. Previously-proposed algorithms, including backprojection, filtered backprojection, and the linear method (here shown with two different choices for the regularization parameters) fail in adequately recovering partially occluded scene parts, such as the logo in the background. The proposed factorization method accurately estimates this challenging scene via the visibility term and it additionally recovers surface normals. Slight variations in intensity, for example for the logo between backprojection and filtered backprojection, are introduced by the volume rendered used to generate these visualizations.}
		\label{fig:sim:dragonplane}
\end{figure*}

In this section, we evaluate the proposed factorization method in simulation and show detailed comparisons to other non-line-of-sight reconstruction methods. Figure~\ref{fig:sim:dragonplane} shows maximum intensity projections generated with the Chimera volume renderer\footnote{\url{http://www.cgl.ucsf.edu/chimera/}} for one of the scenes we used for this evaluation. Several additional results can be found in the Supplemental Material. This scene contains two hidden objects that partly occlude each other (row~1, column~1). The measured transient image is simulated for $64 \times 64$ sampling locations $x',y'$ over an area of 1~m $\times$ 1~m on the visible wall with a temporal bin size of 16~ps (row~1, column~2). A single $xt$-slice of these measurements (row~1, column~3) shows the contributions of time-resolved global illumination effects, i.e. the third bounce of the interreflections from light source to visible wall, to the hidden objects, back to the visible wall, and all the way back to the detector. Direct reflections from the wall are omitted in these visualizations. 

Rows 2--4 of Figure~\ref{fig:sim:dragonplane} show three different perspectives of the reconstructions obtained with the following algorithms: backprojection, filtered backprojection, the linear method with the regularization parameters listed by Heide et al.~\shortcite{Heide:2014:diffusemirrors}, the same linear method with regularization parameters matching those used in the proposed factorization method, the proposed factorization method, and a reference solution obtained by fixing the ground truth visibility term and applying the linear method to recover only the hidden albedos. The latter represents an upper bound on the reconstruction quality that can be achieved with the full factorization method, where the visibility is unknown and also needs to be estimated. Whereas previously-proposed algorithms fail in recovering the partially occluded scene parts, our factorization accurately estimates this challenging scene. Our solution closely matches the reference solution. 

In addition to the hidden albedos, the proposed factorization method also recovers a visibility term (row~1, columns 4--5) and the surface normals of the scene (row~1, column~6). The estimated visibility term contains a full 3D volume of values for each of the $64 \times 64$ sampling locations on the wall. We show one of these volumes for the sampling point indicated by the blue circle (row~1, column~5) as well as the average visibility term for all sampling locations (row~1, column~4). The estimated visibility terms can be interpreted as an intermediate variable that helps improve the estimated albedos in the presence of partial occlusions.  

\begin{table}[t]
\centering
\footnotesize
\begin{tabular}{lllll|l}
\hline
							& BP 	& FBP	& Lin 	& Factored 	& Lin w/ $\mvis$	\\
\hline
$\!\!\!\!$Bunny 							& 15.8 & 13.3 & 26.6 & \textbf{40.8} 	& 46.1 \\
$\!\!\!\!$Dragon 							& 14.4 & 8.8 & 17.6 	& \textbf{19.5} & 19.8 \\
$\!\!\!\!$Dragon \& Logo 			& 14.5 & 10.3 & 23.7 	& \textbf{38.7} & 42.3 \\
$\!\!\!\!$Dragon \& Bunny 		& 14.3 & 10.3 & 22.1 	& \textbf{31.6} & 37.0 \\
$\!\!\!\!$Logo 								& 20.9 & 13.2 & 41.1 	& \textbf{57.9} & 62.2 \\
$\!\!\!\!$2 Planes 						& 19.7 & 13.4 & 29.9 	& \textbf{40.5} & 40.6 \\
$\!\!\!\!$Plane \& Logo 			& 19.9 & 12.1 & 33.3 	& \textbf{48.2} & 49.8 \\
\hline
\hline
							& BP 	& FBP	& Lin 	& Factored 	& Lin w/ $\mnorm$+$\mvis$ \\							
\hline
$\!\!\!\!$Bunny 							& 13.5 & 11.1 & 25.2 	& \textbf{34.4} & 41.9 \\
$\!\!\!\!$Dragon 							& 13.1 & 8.3 & 17.4 	& \textbf{18.7} & 19.7 \\
$\!\!\!\!$Dragon \& Logo 			& 13.6 & 9.4 & 24.3 	& \textbf{30.9} & 46.3 \\
$\!\!\!\!$Dragon \& Bunny 		& 13.3 & 9.5 & 22.3 	& \textbf{27.3} & 40.1 \\
$\!\!\!\!$Logo 								& 17.8 & 9.5 & 39.1 	& \textbf{52.3} & 63.9 \\
$\!\!\!\!$2 Planes 						& 16.4 & 11.0 & 28.0 	& \textbf{37.6} & 53.1 \\
$\!\!\!\!$Plane \& Logo 			& 16.1 & 9.3 & 31.5 	& \textbf{43.4} & 60.9 \\
$\!\!\!\!$Spheres 						& 14.8 & 10.1 & 24.5 	& \textbf{28.5} & 34.9 \\
\hline
\end{tabular}
\caption{Quantitative comparison of various NLOS reconstruction algorithms: backprojection (BP), filtered backprojection (FBP), linear (Lin), and the proposed factorization method. We compare these algorithms for scenes with isotropic BRDFs (top 7 scenes) and scenes with Lambertian BRDFs, including surface normals (bottom 8 scenes). As a reference solution, we also apply the linear method with fixed ground truth visibility term (top) and visibility as well as normals (bottom); these values represent an upper bound on what quality can be achieved. All values are reported as peak signal-to-noise ratio (PSNR) in dB.}
\label{tbl:quantitative} 
\end{table}

Table~\ref{tbl:quantitative} shows quantitative comparisons of these reconstruction algorithms. For this purpose, we simulate a set of scenes with isotropic BRDFs (Tab.~\ref{tbl:quantitative}, top part) without surface normals and a set of scenes with Lambertian BRDFs that include surface normals (Tab.~\ref{tbl:quantitative}, bottom part). We quantify reconstruction fidelity using the peak-signal-to-noise ratio (PSNR). In all cases, the proposed factorization approach results in the highest PSNR. Note that the linear method (Lin) in this experiment uses the same regularization parameters as the linear subroutine in the factored method. The solutions Lin w/ $\mvis$ and Lin w/ $\mnorm$+$\mvis$ apply the linear method with ground truth visibility and ground truth visibility and normals fixed. Due to the fact that the resulting problems are convex, these values can be interpreted as the reference solution representing an upper bound on what PSNR could be achieved. 

\section{Validation with Prototype}
\label{sec:results}
\subsection{SPAD-based Imaging System}

Our prototype uses a single photon avalanche diode (SPAD) and a pulsed picosecond laser. We summarize the hardware components, calibration procedure, and acquisition parameters in the following. 

\begin{figure}[t]
	\centering
		\includegraphics[width=0.9\columnwidth]{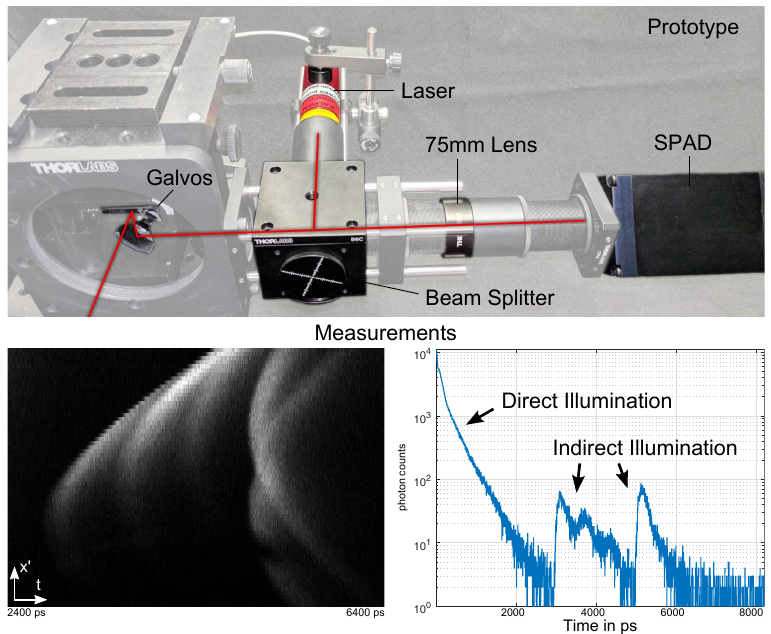}
		\caption{Prototype and measurements. The prototype (top) consists of a single-photon avalanche diode (SPAD), a pulsed laser, a 2-axis scanning galvanometer, and a beam splitter that combines the optical path of the laser and the SPAD (red line). With this setup, we scan a 2D array of sampling locations capturing a temporal histogram of photon counts (lower right) at each location. A spatio-temporal slice of these measurements is shown (lower left), with only indirect illumination inside the displayed area.}
		\label{fig:prototype}
\end{figure}

\begin{figure*}[t!]
	\centering
		\includegraphics[width=\textwidth]{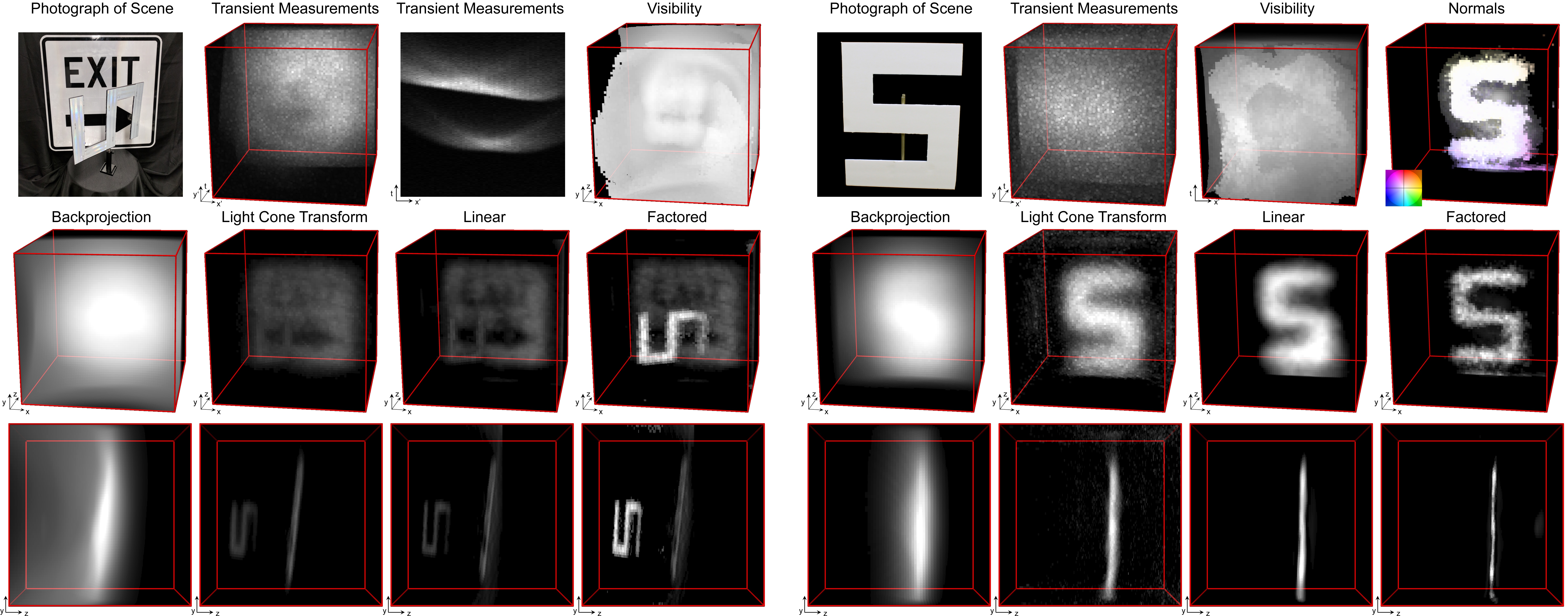}
		\caption{Experimental results of the ``Sign \& S'' (left) and the ``Diffuse S'' (right) scenes. The left scene contains two retroreflective objects that partly occlude each other. We show measurements and compare the reconstruction quality of the backprojection method, the linear method, the Light Cone Transform, and the proposed factorization method. For this example, the proposed method achieves the highest image quality. In the absence of partial occlusions in the hidden scene, all methods, except for the backprojection, achieve a comparable quality. Unlike other methods, the factorization also estimates surface normals of the hidden scene (top right).}
		\label{fig:results}
\end{figure*}

\subsubsection{Hardware}

Our prototype uses a Micro Photon Devices SPAD with a $100 \times 100$~$\mu$m active area. Light is focused on the detector using a 75~mm achromatic doublet lens (Thorlabs AC254-075-A-ML). Photon arrival times are time-stamped with a PicoHarp 300~Time-Correlated Single Photon Counting (TCSPC) module and stored in histograms of photon counts with a 4~ps bin width. The laser is an ALPHALAS PICOPOWER-LD-670-50, which operates at 670~nm wavelength pulsed with a reported pulse width of 30.6~ps at a 10~MHz repetition rate and 0.11~mW average power. The laser and SPAD are co-axially aligned using a polarizing beam splitter cube (Thorlabs PBS251). The aligned optical path is then scanned over the visible wall using a 2-axis scanning galvanometer (Thorlabs GVS012) from a distance of about 1.5~m from the wall. The combined temporal jitter of SPAD and laser pulse width is measured to be approx. 60~ps. An illustration of the setup and measurements are shown in Figure~\ref{fig:prototype}.

\subsubsection{Calibration}

Aligning the laser and SPAD is done by adjusting the beam splitter position and tilt angle to maximize the recorded photon counts of the light directly reflected off the wall. The SPAD is operated in free-running mode, which creates an effect known as \emph{pileup}. Pileup is basically a masking effect that makes it difficult to see weak signals that occur right after a strong signal in the temporal histogram. To avoid masking the weak indirect reflections with the strong contribution of the direct reflections, we slightly misalign SPAD and laser by moving the beam splitter until we can see both direct and indirect contributions (cf. Fig.~\ref{fig:prototype}, lower right). The confocal image formation model is not affected by this procedure. Alternatively, a temporal gating mechanism could be employed to remove the direct light; we did not have access to hardware with this capability for our experiments. 

To account for the differences in path length of different samples on the wall and the imaging system, we align the measured histograms in software such that the peak of the direct light appears at time $t=0$. Then, we reduce the histograms to a bin size of 16~ps using area downsampling and remove the direct light for further processing by setting the first 600 time bins to 0.
 
\subsubsection{Acquisition and Reconstruction Parameters}

In total we captured six scenes. To keep the acquisition times manageable, five of these scenes contain retroreflective objects and only one scene contains a diffuse object. These scenes are recorded with $64 \times 64$ sample points spaced as an equidistant grid on a visible, white planar surface. The sampling points of the retroreflective scenes cover an area of $80 \times 80$~cm of the visible wall; the exposure time for each of the samples is 0.1~s; reconstructed volumes of hidden surface albedos have a resolution of $64 \times 64 \times 120$ voxels and cover $80 \times 80 \times 80$~cm. The diffuse scene was sampled over an area of $70 \times 70$~cm with an exposure time of 1~s per sample. 

For the reconstructions with the linear method, we run 150 iterations with a weight of 0.1 on the sparsity prior and 0.001 on the total variation (TV) prior. The proposed factorization method uses 5~ADMM iterations in total and 20~iterations for the linear method in each of the ADMM iterations. Our source code is implemented with unoptimized MATLAB code and takes about 2~h per scene on a server with an Intel Xeon E5-4620 (2.20~GHz) and 768~GB RAM.


\subsection{Experimental Results}

Figure~\ref{fig:results} shows two of the experimental data sets; all six are shown in the Supplemental Material. We show 3D maximum intensity projections of the acquired measurements and photographs of the imaged objects along with two different perspectives of the reconstructions obtained with several different methods. As expected from our simulations, the backprojection method gives a rough idea of the shape of hidden objects but fine geometric detail is missing. The linear method achieves significantly better results. The Light Cone Transform (LCT) was recently proposed as a computationally very efficient method for NLOS imaging~\cite{OToole:2018}, but it uses similar assumptions as the linear method and achieves a comparable reconstruction quality. The factorization approach recovers this scene exhibiting partial occlusions most accurately among these methods. For the ``Diffuse S'', our factorization method achieves a slightly better quality than the other methods, mainly due to the regularization parameters. Even though these are the same as for the linear method, the number of iterations is different and the additional visibility term result in slight differences in the computed reconstructions. The primary benefit of the proposed method for NLOS scenes containing isolated objects without substantial amounts of occlusions is that we can estimate surface normals. As expected for this planar scene, the estimated normals mostly point towards the scanned wall. 

For the retroreflective ``Sign \& S'' scene, the proposed factorization method amplifies the shape of the ``S'' compared to other reconstruction algorithms. This is mostly due to the fact that the image formation model of the other methods does not adequately model partial occlusions in the hidden scenes, but our method does and thus results in a more accurate reconstruction. A slight mismatch between the assumed $1/r^2$ distance falloff (cf. Eq.~\ref{eq:math:cnlos}) for these retroreflective objects and their actual BRDF likely contributes to the observed intensity boosting. Finally, there is also ambiguity in the factored light transport representation. An object with very low reflectivity (i.e., small albedo) that is unobstructed could produce the same measurements as the same object with a larger albedo but appropriately down-scaled visibility terms. This ambiguity may boost the brightness of certain objects in the recovered volumes, as observed in Figures~\ref{fig:teaser} and~\ref{fig:results} (left). This boosting effect could potentially be mitigated by placing additional regularizers on the visibility terms. However, such an approach would further increase the high computational cost of our method and was not implemented.

\section{Discussion}
\label{sec:discussion}
In summary, we propose a novel light transport representation for non-line-of-sight imaging along with inverse methods that factor the global illumination components of a transient image of a visible surface into a volume of hidden albedos, surface normals, and visibility terms. The visibility terms model partial occlusions in the hidden scene parts. Our simulations indicate that the proposed factorization approach has the potential to improve the robustness and quality of NLOS imaging for complex scenes that often include partial occlusions. With experimentally-captured data we show similar trends, but also reveal that the non-convex nature of the factored light transport model can result in ambiguities in the reconstructions. Overall, the quality of our reconstructions are demonstrated to be better than those obtained by other methods and our method also facilitates surface normal estimation.

One of the biggest challenges of the factored light transport model is the computational cost. Matrix-free implementations of backprojection-type methods require $O(N^3)$ memory whereas an implementation that uses a sparse matrix representation of the light transport requires $O(N^5)$ memory. The linear method has similar memory requirements but a significantly increased computational cost. The Light Cone Transform provides a closed-form solution to NLOS imaging that achieves a similar quality as the linear method with only $O(N^3 \textrm{log} N)$ memory. The computational cost of the proposed factorization is slightly higher than the linear method, but it is in the same order of magnitude. However, the fact that the factored light transport representation requires storing the visibility terms, which are generally not sparse, results in memory requirements of $O(N^5)$ making our method two orders of magnitude more memory demanding than other matrix-free NLOS algorithms.

\subsection{Future Work}

In the future, we would like to replace our laser with a more powerful option to reduce acquisition times and further improve the quality of hidden objects with Lambertian BRDFs. Even though we demonstrate many benefits of the proposed factored light transport, resolving ambiguities of this non-convex formulation should be explored in more detail in future work. Note that such non-convex behavior is not a ``bug'' in our code but a topic beyond the scope of this paper. For example, surface normals could help disambiguate the shape of reconstructed objects; more advanced priors could be placed on the visibility terms; other types of cross-regularizers between visibility, normals, and albedos could also be investigated. Developing new inverse methods or learning them with a data-driven approach are interesting directions for future research. Similar to other NLOS imaging approaches, we assume that only third-order light bounces contribute to the image formation. This assumption could be lifted in future work to account for diffuse interreflections and higher-order light transport effects in hidden scene parts.

\subsection{Conclusion}

Non-line-of-sight imaging is a promising technology that has the potential to unlock unprecedented imaging modalities for a variety of applications. Recent advances in single-photon detector technology and large-scale inverse light transport algorithms have demonstrated that NLOS imaging is feasible in certain conditions. With the proposed methods, we lift several important restrictions of previous algorithms and take steps towards making NLOS imaging more robust. Yet, further research and development is needed to enable NLOS imaging ``in the wild'', i.e. with strong ambient light, at fast acquisition rates, and for more complex scenarios. 



\end{document}